\documentclass{article}

\PassOptionsToPackage{numbers, compress}{natbib}

     \usepackage[preprint]{paper}

\usepackage[utf8]{inputenc} 
\usepackage[T1]{fontenc}    
\usepackage{hyperref}       
\usepackage{url}            
\usepackage{booktabs}       
\usepackage{amsfonts}       
\usepackage{nicefrac}       
\usepackage{microtype}      
\usepackage{color}
\usepackage{booktabs}
\usepackage{graphicx}
\usepackage{booktabs}
\usepackage{multirow}
\usepackage{amsmath}

\newcommand{\blue}[1]{{\color{blue} #1}}
\DeclareMathOperator*{\minimize}{minimize}

\newcommand{\beginsupplement}{%
        \setcounter{table}{0}
        \renewcommand{\thetable}{S\arabic{table}}%
        \setcounter{figure}{0}
        \renewcommand{\thefigure}{S\arabic{figure}}%
}

\title{Towards Improving Spatiotemporal Action Recognition in Videos \\
} 

\author{Shentong Mo, ~~Xiaoqing Tan, ~~Jingfei Xia, ~~Pinxu Ren \\
Carnegie Mellon University\\
\texttt{\{shentonm, xiaoqint, jingfeix, pren\}@andrew.cmu.edu}
}
 
\begin{document}

\maketitle

\begin{abstract}
Spatiotemporal action recognition deals with locating and classifying actions in videos. Motivated by the latest state-of-the-art real-time object detector You Only Watch Once (YOWO), we aim to modify its structure to increase action detection precision and reduce computational time. Specifically, we propose four novel approaches in attempts to improve YOWO and address the imbalanced class issue in videos by modifying the loss function. We consider two moderate-sized datasets to apply our modification of YOWO - the popular Joint-annotated Human Motion Data Base (J-HMDB-21) and a private dataset of restaurant video footage provided by a Carnegie Mellon University-based startup, \blue{\href{https://www.agot.ai/}{Agot.AI}}. The latter involves fast-moving actions with small objects as well as unbalanced data classes, making the task of action localization more challenging. We implement our proposed methods in the GitHub repository\footnote{https://github.com/stoneMo/YOWOv2}.

\end{abstract}


\section{Introduction}




Object detection is one of the most classical problems in the field of computer vision which focuses on detecting \textit{where} and \textit{what} objects are in a given image - specifically the problem of object localization and object classification. Traditional convolutional neural networks (CNNs) see heavy use for recognizing and classifying objects but fail to localize objects in the image. Besides, classification for images containing more than one object has been a great challenge for traditional CNNs. A more significant gain is obtained with the introduction of Region-based CNN (R-CNN) \citep{girshick2014rich}, bringing CNNs into the field of object detection. Under region proposal based framework, R-CNN conducts CNN based deep feature extraction and combines classification and bounding box regression into a multi-task leaning manner. This two-stage object detector, however, suffers from a high computational time and impractical in reality. You Only Look Once (YOLO) architecture \citep{redmon2016you} was later introduced, which was the first end-to-end object detection network. YOLO ``only looks once" at the input image and a single CNN simultaneously predicts multiple bounding boxes and their corresponding class probabilities. It enjoys high accuracy while being surprisingly fast in real-time. On the other hand, anchor-free approaches such as CenterNet \citep{zhou2019objects} detect objects as points by first modeling the center of an object and then uses this predicted center to get the bounding box coordinates. It successfully beats YOLOv3 for 4\% accuracy at the same speed on COCO dataset \citep{zhou2019objects}.

Recently, much attention has been drawn to the field of spatiotemporal action recognition in videos. Videos, unlike images, are time series of images that consist of both spatial components and temporal components. Hence, spatiotemporal action recognition requires both spatial information from the key frame as well as temporal information from the previous frame. You Only Watch Once (YOWO) \citep{kopuklu2019yowo} is the first single-stage CNN architecture designed for real-time spatiotemporal action localization in video streams. Combining a 2D-CNN branch for spatial information with a 3D-CNN branch for spatiotemporal information, YOWO extracts both motion and appearance features and pass them into a channel fusion and attention mechanism for feature aggregation. YOWO is currently the fastest state-
of-the-art architecture on spatiotemporal action localization task. 

We consider two moderate-sized datasets for our project. One is the public Joint-annotated Human Motion Data Base (J-HMDB-21) \citep{Jhuang:ICCV:2013}, which is also used in the YOWO paper \cite{kopuklu2019yowo}. Another is a private dataset of restaurant videos supported by a recent Carnegie Mellon University-based startup, \blue{\href{https://www.agot.ai/}{Agot.AI}}. This dataset consists of annotated footage of workers preparing food at a fast, carry-out style Sushi restaurant. Unlike the standardized dataset such as J-HMDB-21, the Agot dataset has fast-moving actions with small bounding boxes and may suffer from poor class balance and imperfect bounding boxes. We aim to use spatiotemporal action recognition techniques to automate routine tasks such as checkout, inventory management, and quality insurance in restaurants. 

An important characteristic of YOWO is that it has a very flexible architecture and easy to modify. The 2D-CNN and 3D-CNN branches in YOWO could be redesigned and replaced by any arbitrary CNN architectures. In this project, we propose to modify the YOWO architecture to further increase action detection accuracy and increase computational speed as well as modifying the classification loss to deal with imbalanced object classes. We propose four novel methods accordingly. The first approach is called Linknet, which adds linkage between 2D-CNNs and 3D-CNNs in YOWO. The second approach is DIYAnchorBox, which predefines anchor boxes with K-means clustering to stabilize detecting small objects during the training process as motivated by YOLOv2 \citep{redmon2016yolo9000}. The third approach is a two-stage anchor-free implementation called Center3D as motivated by \citep{zhou2019objects}. And the last approach ENFLoss aims to modify the loss function by taking into account the number of frames for different actions.




\section{Related works}

\subsection{Anchor-based approaches}

\textbf{R-CNN} \citep{girshick2014rich} was one of the most important and pioneering work to introduce CNN into the field of object detection. It employs Selective Search \citep{uijlings2013selective} to generate potential bounding boxes in an image and then run a classifier on these proposed boxes. \textbf{Fast R-CNN} \citep{girshick2015fast} and \textbf{faster R-CNN} \citep{ren2015faster} were later introduced to reduce the computation time by sharing computation and using neural networks to propose regions instead of Selective Search. These three algorithms all fall into the category of two-stage object detectors that first apply a Region Proposal Network to generate regions of interests and then pipe those region proposals for object classification and bounding-box regression. Such algorithms enjoy high accuracy rates, however, they are typically slow in practice.

\textbf{YOLO:} \citep{redmon2016look} The need for a two-stage network was eliminated by the YOLO architecture \citep{redmon2016you}, which was the first end-to-end object detector. YOLO treats object detection as a simple regression problem by taking an input image and learning its class probabilities and bounding box coordinates. It was surprisingly fast and produced competitive results in real-time. However, as YOLO processes image globally, it could be difficult for YOLO to localize small objects precisely, such as localizing a flock of birds. 

\textbf{YOLOv2:} \citep{redmon2016yolo9000} Based on YOLO, YOLOv2 uses anchor boxes to predict bounding boxes for increasing recall.  To increase the accuracy of small objects, a passthrough layer from an earlier layer at 26$\times$26 resolution is applied to extract fine-grained features. YOLOv2 also considers multi-scale training to increase the robustness of the model to images of different sizes. To increase the speed of object detection, YOLOv2 applies the global average pooling and the 1$\times$1 filter to compress the feature representation to the Darknet-19 network for decreasing the floating-point operations. 

\textbf{YOLOv3:} \citep{redmon2018yolov3} Compared to YOLOv2, YOLOv3 proposes the Darknet-53 architecture to increase the detection accuracy further. YOLOv3 also uses the independent logistic classifier for overlapping labels and predicts bounding boxes at three different scales for small objects. The Darknet-53 network mainly uses the same successive 3$\times$3 and 1$\times$1 convolutional layers as Darknet-19 but has some shortcut connections to learn much compressive feature representations.

\textbf{3D-CNNs:} For a very long time performance on 3D-CNNs had been much poorer than that on 2D-CNNs because of the small data-scale of video datasets that were available for optimizing the immense number of parameters in 3D-CNNs. Besides, 3D-CNNs can only be trained on video datasets whereas 2D-CNNs can be pre-trained on ImageNet. With the introduction of the Kinetics dataset \citep{kay2017kinetics}, both shallow and deep networks such as ResNet and DenseNet trained on the Kinetics dataset have better performance than 2D-CNNs and complicated 3D-CNNs. Additionally, simple networks pre-trained on the Kinetics dataset outperform complex 2D-CNNs \citep{hara2018can}. 

\textbf{YOWO:} Motivated by the excellent performance of 3D-CNN on video data, YOWO uses a 3D-CNN branch to model the spatiotemporal features of the clip consisting of previous frames. For detecting the location of the action object in videos, YOWO also applies a 2D-CNN branch to extract the spatial features of the key frame. The features from the above two branches are aggregated together using a channel fusion and attention module. In the end, two convolution layers are applied for classification and bounding box regression. So far, YOWO is the first real-time single-stage framework for spatiotemporal action localization. It achieves 3.3\% frame-mAP improvement on J-HMDB-21 dataset compared with previous two-stage methods. YOWO is also the fastest state-of-the-art model: 34 fps for 16-frame clips and 62 fps for 8-frame clips.

\subsection{Anchor-free approaches}

Anchor-free approaches are becoming increasingly popular in that, unlike most successful object detectors we discuss above, they don't need to exhaust all potential object locations in an image and do classification for each location. 

\textbf{CenterNet} \citep{zhou2019objects} detects objects as points in an image. It focuses on modelling the center of a box as an object and then uses this predicted center to get the coordinates of the bounding box. As well as being end-to-end differentiable and simpler, CenterNet has surprising improvements in both detecting speed and accuracy, especially compared to YOLOv3. Under the same speed, CenterNet beats YOLOv3 for about 4\% more accuracy \citep{zhou2019objects}. Some possible challenges such as points collision may exist, however, it didn't affect that much to the accuracy results.

\textbf{CenterTrack} \citep{zhou2020tracking} has been developed to track objects from frame to frame. They use displacement of a point between adjacent frames to track the object in time. The structure is similar to CenterNet. The only difference is that there are two more input: images and heatmap from the frame before and one more output: the displacement of a point.

\textbf{RepPoints} \citep{RepPointsv1} is an anchor free object detector that uses a set of points to learn automatically the spatial extent of an object. This set of points can indicate semantically significant local areas in an image. The whole network is learned via weak localization supervision from ground-truth bounding boxes and implicit recognition feedback. The reason why we use a set of points to represent an object instead of a bounding box is due to the fact that the bounding box provides only a coarse localization information of objects and leads to coarse extracted object features. Motivated by their work, we are planning to design an anchor free detector for spatiotemporal action localization tasks. To the best of our knowledge, this detector will be the first anchor free detector for action localization problems.

\subsection{Resolving class imbalance}
For special data sets with imbalanced object category distribution, object detectors face more difficulties. The main reason for this limitation is the widespread use of cross-entropy loss in object classification tasks. It fail to consider object class frequency during the training process, resulting in a poor performance for objects with a small number of samples.

The imbalance of object detector training can occur in two fashions: one is foreground-objects-to-background imbalance and another one is foreground-object-to-foreground-object imbalance \citep{phan2020resolving}. Foreground-objects-to-background imbalance is inevitable in object detection task since most portion of the image is background. Hence, typically more focus is on solving the problem of foreground-object-to-foreground-object imbalance by focusing on cost-sensitive approach and simply applying different weighted versions of Cross Entropy classification loss. There are several approaches to modify the loss function by assigning suitable weights to each object class such as weighted cross entropy, balanced cross entropy, inverse class sample number frequency, and reweighting based on effective number of object classes \citep{phan2020resolving}.

\section{Datasets}

We propose to use the public Joint-annotated Human Motion Data Base (J-HMDB-21) \citep{Jhuang:ICCV:2013} and a private dataset of restaurant videos generously supported by CMU-based startup \blue{\href{https://www.agot.ai/}{Agot.AI}}.

\textbf{J-HMDB-21} is a subset of the HMDB-51 dataset \citep{kuehne2011hmdb} and consists of 928 short videos with 21 action categories in daily life. Each video is well trimmed and has a single action instance across all the frames. 

\textbf{Agot dataset} consists of labeled video footage of the same restaurant from different vantage points. The combined 100 minutes of 720p footage has about 36,000 still image frames, of which roughly half have no labeled bounding boxes. 
The dataset is highly imbalanced, with some actions having far fewer examples than others. We aim to deal with the class imbalance issue and rapid actions involving only small objects as motivated by the Agot dataset. 

\section{Evaluation metrics}

We use the common precision and recall as our evaluation metrics. Precision measures the proportion of correct predictions while recall measures the proportion of actual positives that are correctly identified. We also use frame-mean Average Precision (mAP) \citep{everingham2010pascal}, which is a popular metric in the field of spatiotemporal action recognition, following the YOWO paper \citep{kopuklu2019yowo}. Frame-mAP measures the area under the precision-recall curve of detection for each frame. The detection of a frame will be classified as correct if both the intersection-over-union (IoU) with the ground truth is greater than a threshold as well as the action label is correctly predicted. We set an IoU threshold to be 0.5 in classifying whether the prediction is a true positive or a false positive.

\section{Working baseline models}

YOWO is the state-of-the-art end-to-end network on spatiotemporal action localization tasks \citep{kopuklu2019yowo}. The YOLOv3 architecture also can achieve competitive performance on object detection problems \citep{redmon2018yolov3}. Therefore, we choose these two networks as our baseline for improvements. We reproduce baseline results using public code from their work. 
Note that we change the final output layer of both original baselines to match the number of classes from our datasets.

\section{Proposed methods}


\subsection{Linknet: adding linkage between 2D-CNNs and 3D-CNNs in YOWO}

One of our proposed work is motivated by the YOWO architecture \citep{kopuklu2019yowo} for spatiotemporal action localization. YOWO is mainly composed of one 2D-CNN branch and one 3D-CNN branch. The 2D-CNN extracts the spatial features of the current frame while the 3D-CNN branch models the spatiotemporal features of the clip consisting of previous frames. The problem is that these two backbones do not have interaction during feature extraction. However, the spatiotemporal features of previous frames are highly related to the spatial features of the key frame. 

We propose to modify the existing YOWO structure to increase accuracy or reduce computational time. Specifically, we propose to reduce necessary computation by reusing 2D feature extractors and more sparsely performing time-wise computation. This is because there’s very little difference between two consecutive frames of a video.
Besides, we propose to link the 2D-CNN branch and the 3D-CNN branch of the YOWO architecture more compactly by adding several interaction layers between them. In this way, information on the key frame for an input clip could also be integrated into the 3D-CNN backbone, which hopefully could reduce the complexity of the 3D-CNN structure and improve computational speed. 

\begin{figure}[!htp]
    \centering
    \includegraphics[width=0.5\textwidth]{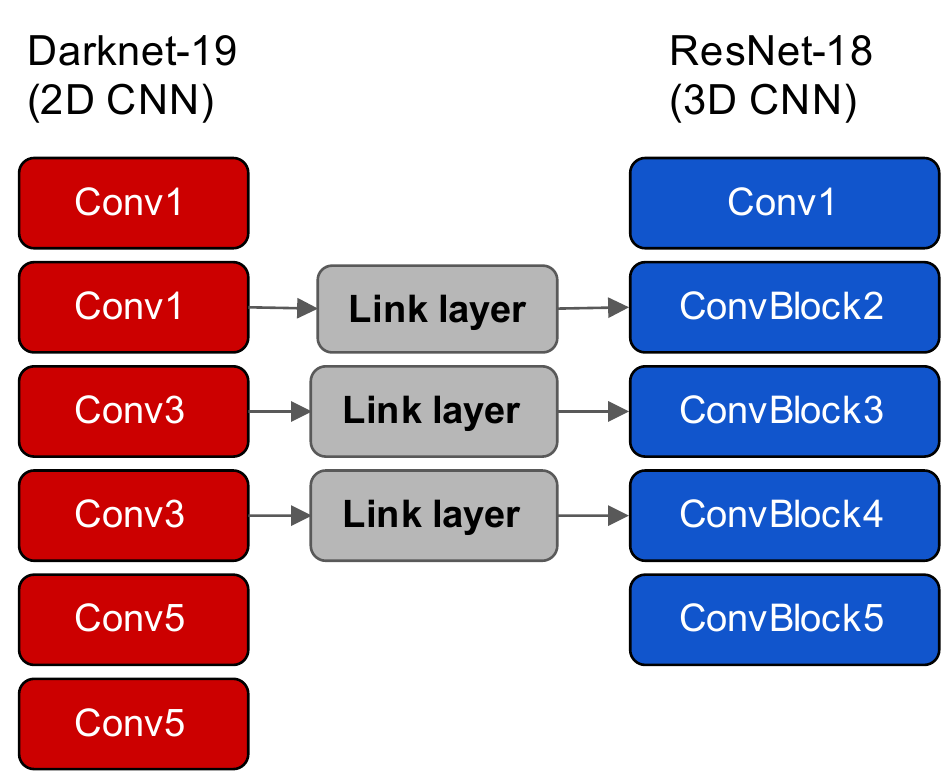}
    \caption{Linknet structure}
    \label{fig:linknet_structure}
\end{figure}

As we have mentioned above the basic idea of Linknet is to connect 2D and 3D networks so that in the 3D network we can use the information generated by the 2D network. In this experiment, we use Darknet-19 and Resnet-18 as our 2D and 3D networks, respectively. Figure \blue{\ref{fig:linknet_structure}} shows the structure of Linknet structure. The block in 2D CNN represents several consecutive CNN layers. The number represents the number of CNN layers. The block of 3D CNN contains several numbers of basic blocks. We can see that we have added three connections between the network. To keep the network parameters the same and use pretrained model for each network, we simply expand the dimension of 2D features in Darknet and add the information to Resnet in the corresponding stage. We have tried different link layers: add 2D feature directly to the 3D network, use one convolution layer, and use bottleneck structure with two convolution layers. The rest of the network is the same as that in YOWO. We compare the performance with the YOWO structure using Darknet-19 and Restnet-18 and see if there is any improvement within this structure.

\subsection{DIYAnchorBox: predefine anchor boxes with K-means clustering}

For detection purposes, multiple ways could be done to compute bounding boxes. A common approach is to predict the coordinates of bounding boxes directly. This, however, may be susceptible to bias as the training process could be unstable, especially in the case of small objects and insufficient training samples. An alternate way is to use pre-defined bounding boxes called anchor boxes before training, and adjust these priors with the ground truth bounding boxes during training \citep{redmon2016yolo9000}. YOLOv2 \citep{redmon2016yolo9000} computes anchor boxes by K-means clustering, achieving a better result than the original YOLO \citep{redmon2016look}, where no anchor boxes are predefined. Following a similar idea, we propose to compute candidate anchor boxes using K-means clustering for the YOWO model. The number of cluster $K$ is selected to minimize the inertia or within-cluster sum-of-squares criterion
\begin{align*}
    \sum_{i=0}^n \displaystyle{\minimize_{\mu_j\in C} ||x_i - \mu_j||^2}
\end{align*}
where $n$ is total number of training samples, $x_i$ is the $i$-th training sample. $C$ denotes clusters, and $\mu_j$ denotes mean of the samples in each cluster.

\subsection{Center3D: two-stage anchor-free implementation}

We try to use an anchor-free structure to localize objects and classify different actions. In our project, instead of using just one point, we plan to use several points to represent the target object. The current CenterNet structure provides a function of pose estimation. It will generate several points on the body of a single person \citep{zhou2019objects}. We will use the points generated by CenterNet to do localization and action classification tasks. For the localization part, we will use CenterNet only. Every time we generate several points of a person, we also generate the offset between those points and the center point of the object. We also generate the width and length of the bounding box. With those features, we can construct a unique bounding box for each object. For the action classification problem, we would create a heatmap for each frame which includes the points information of the object. In this case, each frame would convert to a heatmap containing several points generated by CenterNet. We then train pretrained 3D CNN with these heatmaps as input and output the action label. In this experiment, we use do not train the CenterNet on our dataset. We just use CenterNet trained on the COCO dataset and it generates good performance on our dataset. So in total, we can consider the method as a two-stage method: firstly a CenterNet is used to localize the object and generate a heatmap for each frame, and then a 3D-CNN is used to classify the action in each frame.

\begin{center}
    \includegraphics[width=0.9\textwidth]{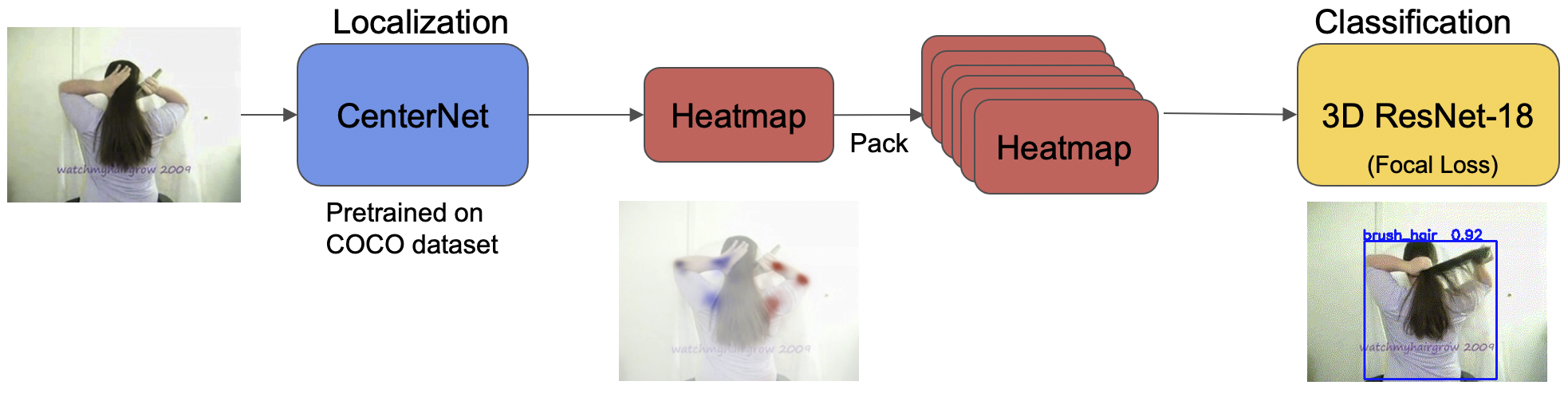}
\end{center}

\subsection{ENFLoss: focal loss with weight combinations}

Motivated by the class imbalance issue in Agot data, we propose to modify the classification loss function by reweighting the focal loss with effective number of object classes as discussed in \citep{phan2020resolving}.




YOWO applies focal loss for classification \citep{kopuklu2019yowo}. Focal Loss uses the confidence of object classifier with a weight factor 
\begin{align*}
w_1 = [1- P(i)] ^{\alpha}
\end{align*}
where $\alpha$ is a real positive hyperparameter and $P(i)$ is the confidence of $i^{th}$ object proposal belonging to each given class given by the object classifier \citep{phan2020resolving}. As the classifier is less confident in minority-class objects, a lower value of $P(i)$ makes the correspondent weight $w$ larger and automatically brings focus to itself in the training of object classifier.

The effective number of object class is another weighting scheme that uses the effective number of samples for each class to re-balance the classification loss. In our project, the effective number of samples is defined as the total frames of an object class in video and the weight of each object class is calculated by
\begin{align*}
w_2 = \frac{1- \beta ^{n_j}}{1- \beta}
\end{align*}
where $n_j$ is the frames of object class in the $j^{th}$ object class with $\beta \in [0,1)$ as a hyperparameter.

Inspired by \citep{phan2020resolving}, our implementation of weighted balanced loss function combines both focal Loss and the effective number of object class by multiplying the above-mentioned weight factors as suggested in \citep{phan2020resolving}. Specifically, the updated weight of each class is 
\begin{align*}
w = w_1 * w_2 = [1- P(i)]^{\alpha}\frac{1- \beta ^{n_j}}{1- \beta}
\end{align*}
We first test these hyperparameters $\alpha$ and $\beta$ on small samples to see if they balance the small sample dataset well. The hyperparameters we choose are $\alpha=2$ and $\beta = 0.7$. 



\section{Results}

\subsection{Data preprocessing}

\begin{figure}[htp]
    \centering
	\includegraphics[width=0.75\textwidth]{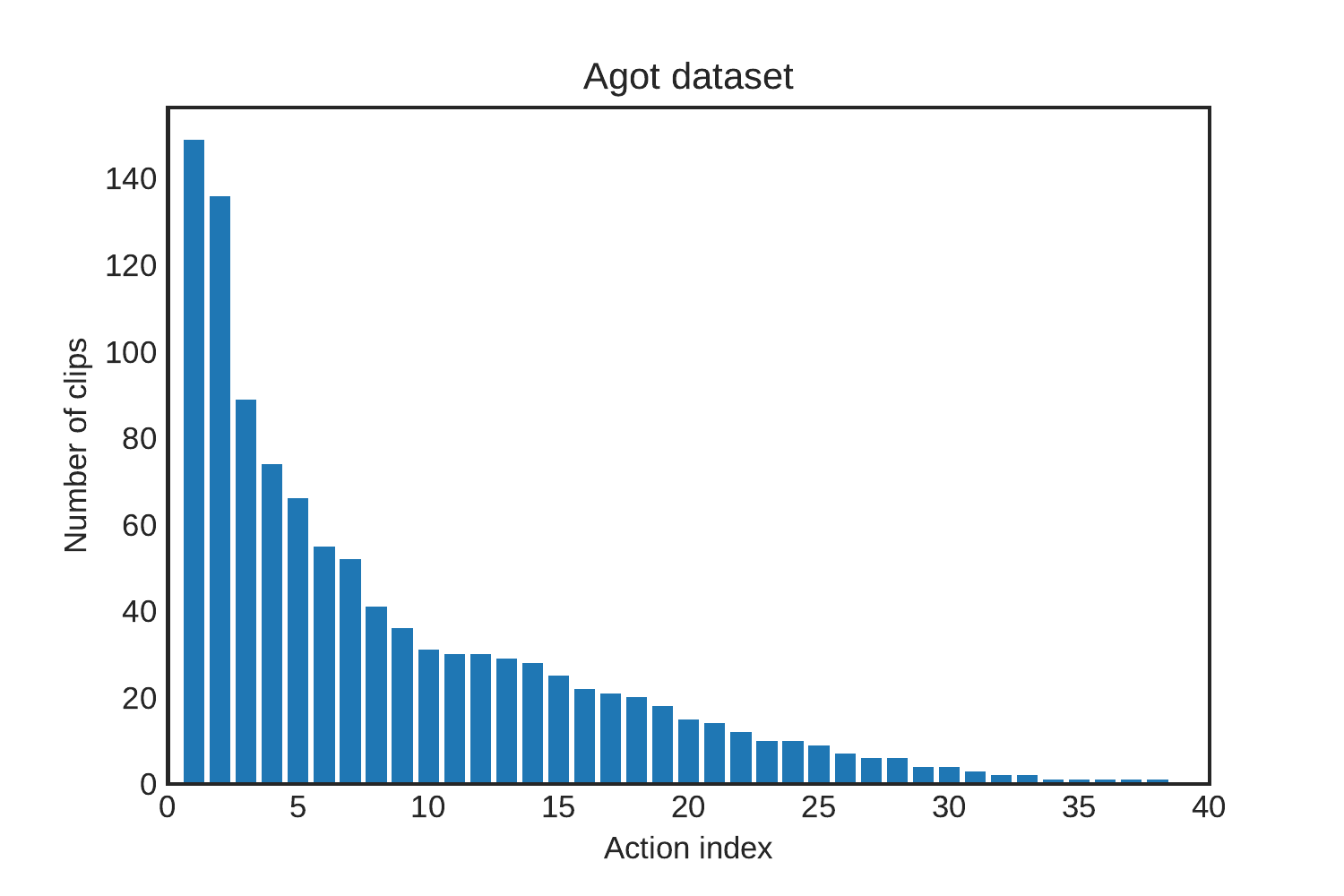}
	\caption{Distribution of actions in the Agot dataset. Each action is ranked by its number of clips in the data. The top 5 actions with the most number of clips) are ``pick up from bin with tong or scooper", ``put item into meal using tongs"", ``put tongs or scooper back in bin", ``operating POS", ``put item into meal using hands".}
	\label{fig:agot_hist}
\end{figure}

Data preprocessing is not needed for the public J-HMDB-21 dataset, however, it is required for the Agot dataset. Among the 45 actions in the Agot dataset, 38 actions are labeled. Figure \blue{\ref{fig:agot_hist}} shows the number of clips per action. As shown in the figure, the dataset is highly imbalanced, with some actions such as ``pick up from bin with tong or scooper"" and ``put item into meal using tongs"" having about 140 clips while others only have less than 90 clips. We select the top 24 actions with 10 or more labeled clips as our dataset for analysis. Table \blue{\ref{tab:agot_top}} shows the number of clips and frames among the 24 classes of actions. We further split the preprocessed Agot data into a training set and a testing set with a proportion of 7:3, making sure the training and testing sets consist of at least each of the 24 actions. 

\begin{table}[!htp]
\caption{Distribution of actions in Agot dataset, ranked by number of video clips.}
\label{tab:agot_top}
\centering
\begin{tabular}{@{}clcc@{}}
\toprule
\multicolumn{1}{l}{Action index} & Action label                          & \multicolumn{1}{l}{Number of clips} & \multicolumn{1}{l}{Number of frames} \\ \midrule
1                                & Pick up from bin with tong or scooper & 149                                 & 7187                                 \\
2                                & Put item into meal using tongs        & 136                                 & 5084                                 \\
3                                & Put tongs or scooper back in bin      & 89                                  & 1015                                 \\
4                                & Operating POS                         & 74                                  & 9414                                 \\
5                                & Put item into meal using hands        & 66                                  & 3234                                 \\
6                                & Pick up from bin with hands           & 55                                  & 2538                                 \\
7                                & Pick up sauce                         & 52                                  & 954                                  \\
8                                & Put sauce down                        & 41                                  & 905                                  \\
9                                & Clean counter                         & 36                                  & 4959                                 \\
10                               & Squirt sauce                          & 31                                  & 2782                                 \\
11                               & Put roll into roll cut machine        & 30                                  & 1327                                 \\
12                               & Place cut roll into boat              & 30                                  & 7734                                 \\
13                               & Pickup cut roll from roll machine     & 29                                  & 4687                                 \\
14                               & Operating sushi roll cutter           & 28                                  & 3215                                 \\
15                               & Roll a sushi roll                     & 25                                  & 10407                                \\
16                               & Pick up a drink                       & 22                                  & 593                                  \\
17                               & Hand item to customer                 & 21                                  & 736                                  \\
18                               & Put gloves on                         & 20                                  & 3195                                 \\
19                               & Handing credit card                   & 18                                  & 514                                  \\
20                               & Put lid on roll                       & 15                                  & 845                                  \\
21                               & Put drink down                        & 14                                  & 403                                  \\
22                               & Operating fivestar                    & 12                                  & 2209                                 \\
23                               & Using a cell phone                    & 10                                  & 15239                                \\
24                               & Inserting chip of credit card         & 10                                  & 838                                  \\ \bottomrule
\end{tabular}
\end{table}

\subsection{Experiment results on J-HMDB-21 dataset}

\begin{figure}[!htp]
    \centering
    \includegraphics[width=\textwidth]{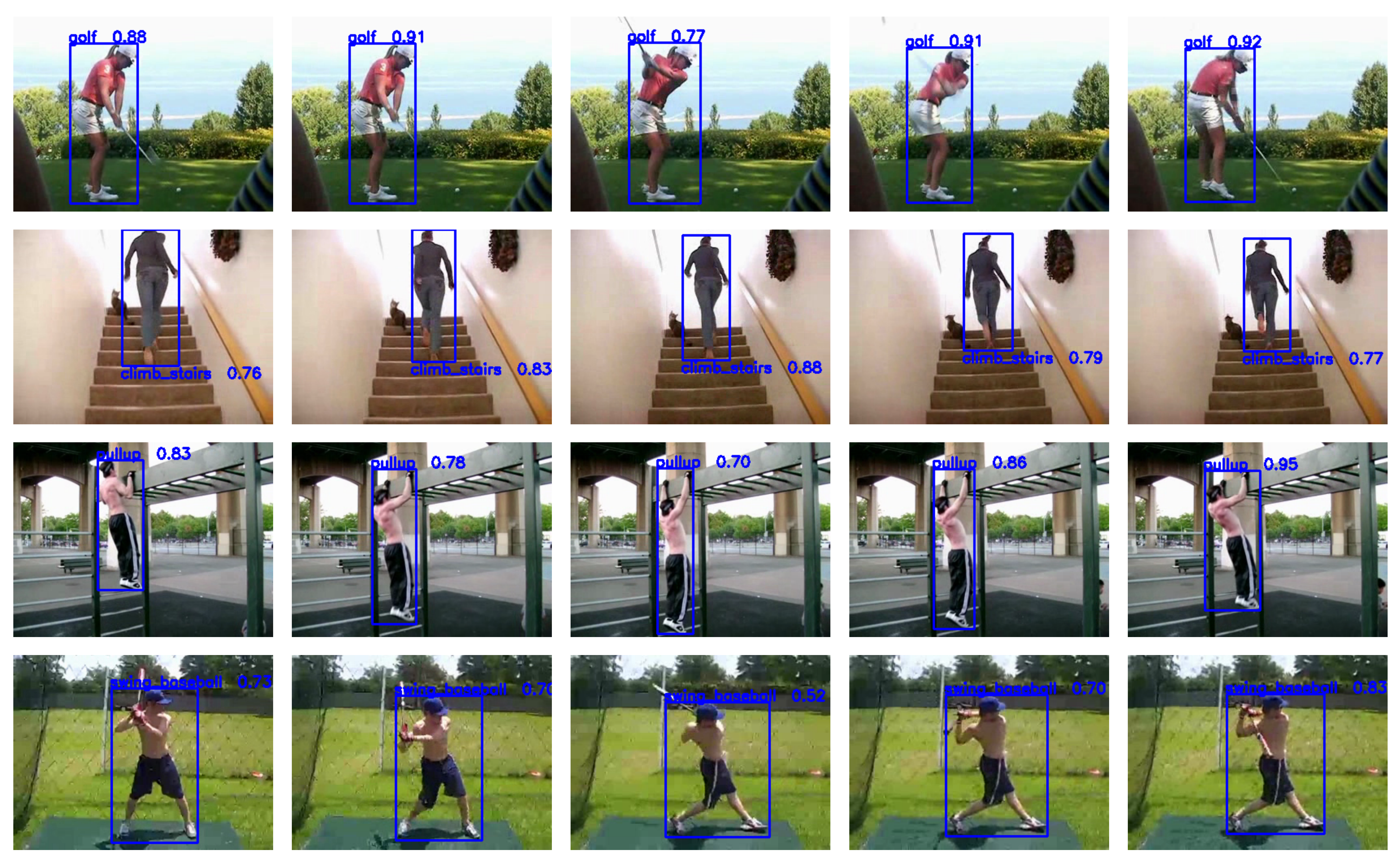}
    \caption{Visualization Results on J-HMDB-21}
    \label{fig:vis_jhmdb}
\end{figure}

The experiment results of using YOWO baseline and our proposed networks Center3D and Linknet with variants are shown in Table \blue{\ref{tab:jhmdb_res}}, respectively. For the sake of time and computational resources, we use Darknet-19 and Resnet-18 for 2D-CNN structure and 3D-CNN structure in all models. We achieve consistent results for the baseline YOWO as in \cite{kopuklu2019yowo}. We show visualization results in Figure \blue{\ref{fig:vis_jhmdb}}.

Our proposed network, Linknet does not perform very well in this situation. Expand means expanding 2D data directly to the shape we want in a 3D network. As we can see from the result, the localization recall does not change a lot but the classification accuracy becomes low after adding the link layers. We can conclude that the 2D features mainly contribute to the localization which 3D features or temporary feature mainly contributes to the action classification. The reason why the result becomes less ideal may be that the 2D features blur the information in 3D. Simply using expansion or several convolution layers will minimize the difference between different frames and thus make it more difficult to classify actions. However, if we continue using more complex structures, the training time would increase and it will make this link structure meaningless. 

The last row in Table \blue{\ref{tab:jhmdb_res}} shows the performance results of our proposed Center3D model on J-HMDB-21 dataset. It outperforms the baseline by a large margin in terms of localization recall and training speed. Its classification accuracy and frame mAP are also comparable with our baseline. The reason why the training speed is faster is that we fix the parameters of the CenterNet, and update the parameters from the 3D CNN part. In terms of the localization recall, the performance of our Center3D model is better than the baseline since we use CenterNet as the network to localize where the action happens. However, the classification accuracy is lower compared to the baseline, since we need to reshape the extracted heatmaps to a small size because of machine resource limits.  

\begin{table}[htp]
\label{tab:jhmdb_res}
\centering
\caption{Performance of different models on J-HMDB-21 dataset. We use Darknet-19 and Resnet-18 as our 2D backbone and 3D backbone, respectively, in all experiments.}
\begin{tabular}{@{}cccccc@{}}
\toprule
\multirow{2}{*}{Model}                                               & 2D-3D                                                    & Training & Localization & Classification & Frame mAP \\
                                                               & Linkage                                                  & (/epoch) & Recall       & Accuracy       & (IoU=0.5) \\ \midrule
Baseline & None         & 0.55h    & \textbf{97.6}         & \textbf{69.6}           & \textbf{68}        \\
Linknet  & expand       & 0.55h    & 97.3         & 54.9           & 53.2      \\
Linknet  & conv3d       & 0.63h    & \textbf{97.6}         & 54.9           & 53.4      \\
Linknet    & \begin{tabular}[c]{@{}c@{}}expand\\ +conv3d\end{tabular} & 0.65h & 97.2 & 53.3 & 51.7 \\
Center3D  & /       & \textbf{0.38h}    & \textbf{99.1}         & 63.1           & 62.7      \\
\bottomrule
\end{tabular}
\end{table}

\subsection{Experiment results on Agot dataset}

\begin{figure}[!htp]
    \centering
    \includegraphics[width=0.9\textwidth]{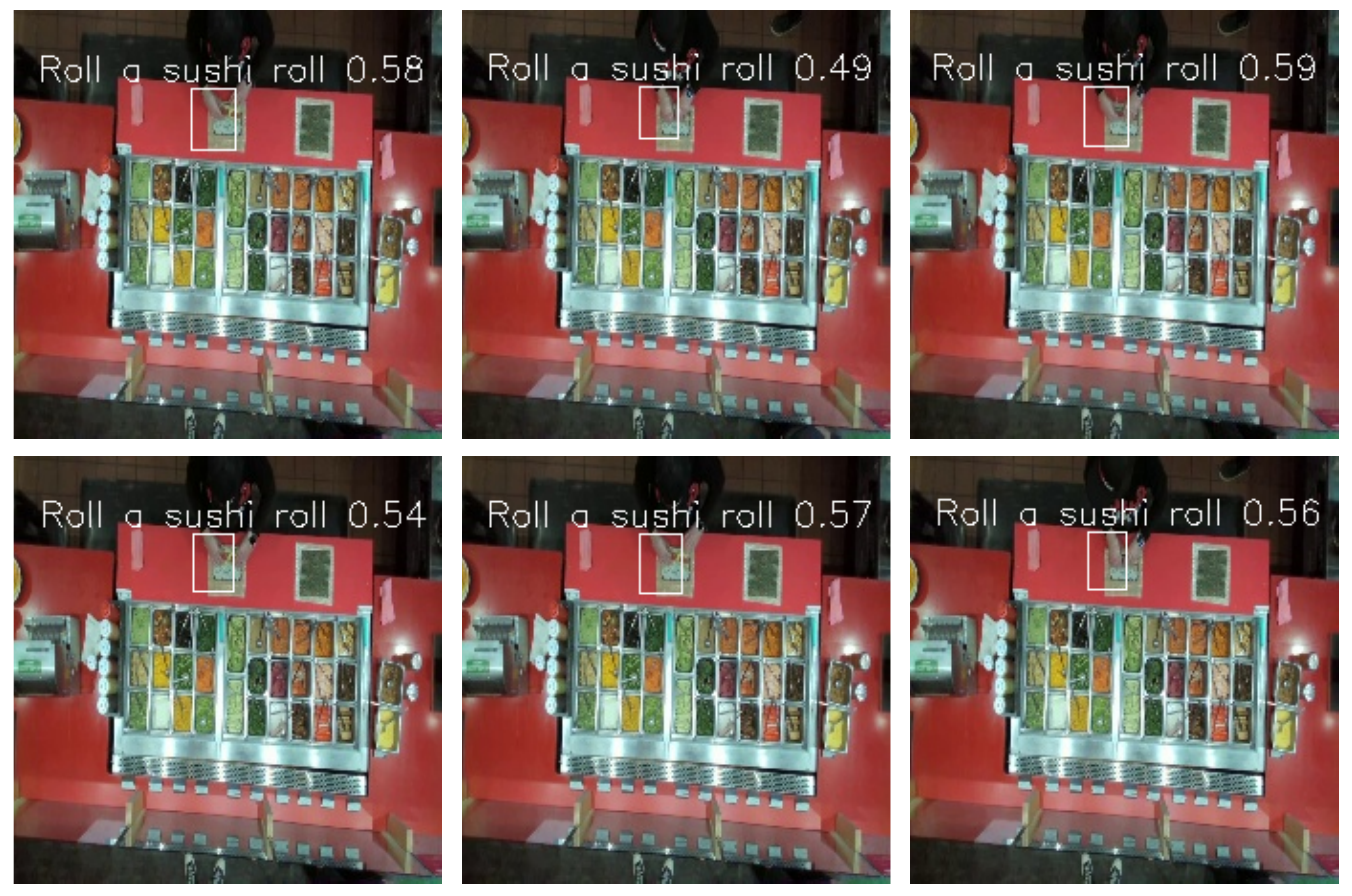}
    \caption{Visualization Results on Agot dataset}
    \label{fig:vis_agot}
\end{figure}

We run the baseline YOWO, YOLOv3 models as well as our proposed methods DIYAnchorBox and ENFLoss on the Agot dataset. The experiment results are shown in Table \blue{\ref{tab:agot_res}}. We decide not to train YOWO with Linknet due to the poor performance of it on the J-HMDB-21 dataset. We show visualization results in Figure \blue{\ref{fig:vis_agot}}.

For DIYAnchorBox, $K=5$ is chosen to be the number of clusters as it minimizes the within-cluster sum-of-square. We achieve a better performance in terms of accuracy and frame mAP compared to the baseline YOWO architecture. This may be because with appropriate anchor boxes selected, the training process becomes more stable when applying to the classification on the Agot dataset with small objects and few training samples. However, DIYAnchorBox fails to give a better localization performance.

Although ENFLoss has a comparable localization recall compared to the YOWO baseline, it suffers from a very low classification accuracy and a low frame mAP. It would be better if there is more fine-tuning on the hyperparameter. 

Meanwhile, YOLOv3 baseline achieves better performance than the YOWO models in terms of localization recall and frame mAP. This is because YOLOv3 can extract more representative spatial features from the key frame. But due to the lack of temporal information from the past frames, YOLOv3 can not achieve comparable classification accuracy as compared to YOWO models. 

\begin{table}[htp]
\centering
\caption{Performance of different models on Agot dataset.}
\label{tab:agot_res}
\begin{tabular}{ccccccc}
\toprule
\multirow{2}{*}{Model} & 2D         & 3D        & Training & Localization & Classification & Frame mAP \\
                       & Backbone   & Backbone  & (/epoch) & Recall       & Accuracy       & (IoU=0.5) \\ 
\midrule
Baseline                   & Darknet-19 & Resnet-18 & 1.2h     & 32.5           & 91             & 8.01      \\
DIYAnchorBox            & Darknet-19 & Resnet-18 & 3h       & 30.9         & \textbf{93.8}           & \textbf{8.39}      \\

ENFLoss      & Darknet-19   & Resnet-18     & 2h   &   32.3      & 84.4    & 5.36\\
\bottomrule
\end{tabular}
\end{table}

Comparisons of mAP between different models for each action in Agot dataset are shown in Table \blue{\ref{tab:map_comp}}. All the four models achieve consistently satisfactory performance in terms of action index 11 (Put roll into roll cut machine), 12 (Place cut roll into boat), 15 (Roll a sushi roll). These actions tend to have a larger number of clips and a larger number of frames, as shown in Table \blue{\ref{tab:agot_top}}. Surprisingly, the YOLOv3 network achieve high frame mAP (86.9\%, 96.6\%)
at action index 22 (Operating fivestar) and 23 (Using a cell phone), while both two YOWO models have very low mAP (1.7\%, 0.7\% / 0.6\%, 2.0\%). This could be because the YOLOv3 can extract better spatial features from the key frame for action localization. 

For the ENFLoss that weighted loss is applied, although it has the lowest frame mAP among other models,  it indeed balanced the frame mAP between different action indexes. Also, at action index 4 (Operating POS), 9 (Clean counter), 22 (Operating fivestar) the ENFLoss have even better results than the YOWO and DIYAnchorBox model. With fine-tuning of hyperparameters, we might be able to achieve a better result.

\begin{table}[htp]
\centering
\caption{mAP (\%) comparison between different models for each action in Agot dataset.}
\label{tab:map_comp}
\begin{tabular}{@{}ccccc@{}}
\toprule
Action index & YOWO          & DIYAnchorBox   & YOLOv3        & ENFLoss  \\ \midrule
1            & 6.4           & 1.2           & 56.7          & 3.34     \\
2            & 3.9           & 3.8           & 48.9          & 4.18     \\
3            & 2.2           & 0.1           & 15.5          & 0.0      \\
4            & \textbf{27.6} & \textbf{27.1} & \textbf{75.0} & \textbf{29.39}    \\
5            & 8.3           & 4.9           & 48.5          & 7.6      \\
6            & 3.0           & 7.0           & 65.0          & 3.41     \\
7            & 0.0           & 0.0           & 25.6          & 0.0      \\
8            & 0.0           & 0.0           & 11.3          & 0.0      \\
9            & 11.1          & 10.3          & 86.2          & 12.61    \\
10           & 2.0           & 1.4           & 58.3          & 0.0      \\
11           & \textbf{19.7} & \textbf{42.1} & \textbf{80.2} & 0.0      \\
12           & \textbf{21.5} & \textbf{19.9} & \textbf{91.5} & \textbf{15.2}     \\
13           & 1.9           & 0.2           & 67.0          & 0.79     \\
14           & 17.9          & 5.3           & \textbf{97.4} & 11.11    \\
15           & \textbf{44.3} & \textbf{39.3} & \textbf{96.2} & \textbf{21.36}    \\
16           & 0.0           & 0.0           & 0.9           & 0.0      \\
17           & 4.4           & 8.3           & 55.4          & 0.0      \\
18           & 1.1           & 0.5           & 26.8          & 1.62     \\
19           & 9.8           & 24.3          & 49.5          & 0.96     \\
20           & 0.2           & 2.9           & 61.9          & 0.0      \\
21           & 0.0           & 0.0           & 0.0           & 0.0      \\
22           & 1.7           & 0.7           & \textbf{86.9} & \textbf{16.74}    \\
23           & 0.6           & 2.0           & \textbf{96.6} & 0.27     \\
24           & 6.8           & 0.0           & 26.0          & 0.0      \\ \bottomrule
\end{tabular}
\end{table}

\section{Discussion and future works}

In summary, we have proposed four novel methods in attempts to improve the state-of-the-art YOWO and address the imbalanced class issue motivated by our data. Among our proposed methods, DIYAnchorBox and Center3D performed well among the proposed methods. So far, we train baseline YOWO and our proposed networks Linknet and Center3D on J-HMDB-21 dataset as well as training baseline YOWO, YOLOv3, and our prosed methods DIYAnchorBox and ENFLoss on Agot dataset. We fail to achieve a better performance using the Linknet compared to the baseline, which could be because the 2D features blur the information in 3D-CNNs. On the other hand, we achieve better results in terms of classification accuracy and frame mAP using DIYAnchorBox compared to baseline YOWO. Prespecification of appropriate anchor boxes may help stabilize the training process especially in the case of small objects and insufficient training samples in Agot dataset. Our anchor-free approach Center3D, on the other hand, have a very good recall in a short training time. However, the classification part does not perform better because we use features to represent the original images, which makes the training much faster. It might be useful if we expand the feature size in future experiments. YOLOv3 is getting even better results in terms of localization recall and frame mAP, which suffers from a very low accuracy compared to those YOWO models. This is because YOLOv3 is very good at extracting representative spatial features from the key frame while fails to perform well in terms of extracting temporal features.

Our project has several limitations. Because of time and resource limits, we failed to run Center3D on Agot data. Also, pre-trained models were used in Center3D. For ENFLoss, more fine-tuning on the hyperparameter is needed to achieve a better result. We may also try combining our proposed methods while modifying the loss function for future works.

\bibliographystyle{unsrt}
\bibliography{paper.bib}

\newpage
\beginsupplement

\section*{Supplementary Material}
\section*{A. Visualization Results}

\end{document}